# Linear Antenna Array with Suppressed Sidelobe and Sideband Levels using Time Modulation

Swaprava Nath (swaprava@yahoo.co.in), Subrata Mitra (subrata4096@yahoo.co.in), students, Jadavpur University, Kolkata 700032

*Abstract*—In this paper, the goal is to achieve an ultra low sidelobe level (SLL) and sideband levels (SBL) of a time modulated linear antenna array. The approach followed here is not to give fixed level of excitation to the elements of an array, but to change it dynamically with time. The excitation levels of the different array elements over time are varied to get the low sidelobe and sideband levels. The mathematics of getting the SLL and SBL furnished in detail and simulation is done using the mathematical results. The excitation pattern over time is optimized using Genetic Algorithm (GA). Since, the amplitudes of the excitations of the elements are varied within a finite limit, results show it gives better sidelobe and sideband suppression compared to previous time modulated arrays with uniform amplitude excitations.

## I. INTRODUCTION

The primary motivation of this paper is to design an antenna array with suppressed sidelobes and sidebands. To achieve this in ordinary antenna arrays, where we give fixed excitation to all elements and these excitations are invariable over time, we need quite high excitation levels. To achieve a sidelobe level (SLL) of –30 dB in ordinary arrays, we need to excite the central element almost 32 times higher than the terminal elements of the array. This sort of array is quite difficult to fabricate and consumes much power.

Whereas, varying the excitation amplitude over time we need the maximum amplitude of a particular element just 8 times higher than the minimum excitation level, to achieve the same –30 dB SLL. Here, we don't give fixed excitation to all the array elements but vary it dynamically with time. It takes account of the fourth dimension—time—into the design. It has more flexibility since we have several options to select the time sequences that can give lower side lobe levels (SLL) and sideband levels (SBL).

However, the inherent drawback of time-modulated arrays is that there are many sideband signals spaced at the multiples of the modulation frequency, which implies that part of the radiated or received power is shifted to the sidebands. So, in our optimization, we suppressed the SBL significantly so that it has little effect on the operation of the array at the center frequency.

In this paper, a novel approach to realize variable amplitude time modulated linear arrays with both suppressed side lobes and sidebands is proposed. This is based on the direct optimization of the time sequence and the excitation levels of the different antenna elements of the array. The optimization takes the help of Genetic Algorithm (GA) [1]-[2]. We use GA due to the multimodality and nonlinearity of the problem. Since we are giving variable excitations to the array elements, for the sake of simplicity, we assume that the excitations vary within a limit with integer values. The time modulation period is divided into numerous minimal time steps of same length. As compared to previous time modulated linear arrays [3] with uniform excitation, this novel array possesses the advantages of lower SLL.

## II. THEORY

Let us consider a broadside linear array [4] of *2N* equally spaced isotropic elements, in which each element's excitation can vary within a certain range and is controlled by a high speed RF switch. The array is symmetric in both geometry and "switch on" excitation levels with respect to the array center. The uniform minimal time step of the RF switches is $\tau$, the time modulation period is $T_p$, and the modulation frequency is $prf = 1/T_p$. Consequently, each of the array elements has a total number of $L = T_p/\tau$ time steps in each modulation period within which they are either OFF or excited to a CERTAIN level. For broadside beams, the array factor is given by [4]-[5]

$$E(\theta,t) = 2e^{j2\pi f_0 t}\sum_{k=1}^{N}U_k(t)\cos\left[(2k-1)\frac{\pi d}{\lambda}\sin\theta\right] \quad (1)$$

Where $f_0$ and $\lambda$ are the center frequency and wavelength, respectively, $\theta$ is the angle measured from broadside, d is the element spacing and $U_k(t)$ is the status of the k[th] antenna element at time t.

The optimization of the time sequences and their excitations are done using GA [1]. If we use a gene $g_q^k$ to represent the status of k[th] antenna element at q[th] time instant (e.g. $g_q^k$=0 when switch is OFF and equal to a certain integer value corresponding to its excitation level), the entire switching time sequences for the N elements numbered from the array center to the array edge can be represented by a chromosome

$\chi$, namely

$$\chi = g_1^1\ g_2^1 \ldots g_L^1\ g_1^2 \ldots g_1^k\ g_2^k \ldots g_L^k \ldots g_L^N \quad (2)$$

By decomposing (1) into a Fourier series, the radiation patterns at each harmonic frequency $F_d = m.prf\ (m=0,\pm1,\pm2\ldots\pm\infty)$ are readily obtained, given by [4]-[5]

$$|E_m(\theta)| = 2\left|\sum_{k=1}^{N} a_{mk} \cos\left[(2k-1)\frac{\pi d}{\lambda}\sin\theta\right]\right| \quad (3)$$

where,

$$a_{mk} = \frac{\sin(\pi m \bar{\tau})}{\pi m} \sum_{q=1}^{L} g_q^k e^{-j\pi m \bar{\tau}(2q-1)} \quad (4)$$

and $\bar{\tau} = \tau/T_p$. Using the GA, the chromosome length is L by N, and the cost function can be selected to suppress the SLL and SBL simultaneously, namely

$$f^{(n)}(\chi) = w_1.SLL^{(n)}(\chi)\big|_{f_0} + w_2.SBL^{(n)}(\chi)\big|_{f_0+m.prf} \quad (5)$$

where (n) is the number of evolution generations, SLL is the sidelobe level at the center frequency $f_0$; SBL is the Side Band Level at selected m sideband frequencies, $w_1$ and $w_2$ are the corresponding weighting factors for each term. Where, we define,

SBL = $\min\{|E_0(0)|_{dB} - |E_n(\theta_1)|_{dB}\}; n \in [1,10]$
SLL = $\min\{|E_0(0)|_{dB} - |E_0(\theta_1)|_{dB}\}$

Such that, $E_n'(\theta_1) = 0, 0 \le \theta_1 \le \frac{\pi}{2}$

### III. OPTIMIZATION PROCEDURE

In the optimization procedure we have used GA or Genetic Algorithm [1]-[2] to maximize the fitness function-

$$f^{(n)}(\chi) = w_1.SLL^{(n)}(\chi)\big|_{f_0} + w_2.SBL^{(n)}(\chi)\big|_{f_0+m.prf}$$

Now to optimize the values of SLL and SBL we need to choose suitable values of the weighting factors $w_1$ and $w_2$ depending on our requirements.

If our prime importance is to improve the SBL then obviously we should choose higher value of weighting factor corresponding to SBL. And here we should remember that the value of SBL we have used in this expression is the minimum SBL in magnitude among 10 sideband frequencies.

For our purpose we have chosen a $\lambda/2$ spaced time modulated linear array of 16 isotropic elements. The target is to optimize the "switch-on" time sequences, suppressing the SLL and SBL simultaneously to be as low as possible. In this case, N=8, a minimal time step $\tau = 1\ \mu s$ is selected, and the modulation period T$_p$=10 $\mu s$ implying a modulation frequency of 100 KHz. Using GA the code length L=10, and chromosome length is 80.

Our Genetic Algorithm method to optimize the excitation pattern for suppressing both SLL and SBL is a little modified version of the actual Simple Genetic Algorithm (SGA) [1]-[2]. Here at first we have initialized all the chromosomes of all the populations in the initial generation by random integer values in the range 0 to 7. Then we have calculated the fitness values of each individual member that is each chromosome in the population according to equation (5) where the mathematical details about the calculation of SLL and SBL is given before. We have used an initial population of 60. To introduce elitism we have stored the best 25 members in the population in a separate or extended portion of the original array containing the members, with the best member at the topmost location in the extended portion of the array. As we have said, we have applied elitism in a little bit unconventional manner. For the reproduction process through which the next generation is created, we have attached a tag to each member in the population to discriminate who are eligible for mating and who are not. To do this we have used a flag called "Survivor" which is "TRUE" if the corresponding member is fit for mating and "FALSE" if it is not. The best 25 members, which were stored in the extended portion of the array, always have their flag Survivor equal to TRUE, i.e. they will always play a role in mating. For the rest of the members, those who have their fitness values greater than or equal to the average fitness value of that generation would have their flag Survivor equal to TRUE, and for others it would be FALSE. We have also calculated the total number of survivors in the variable N Survivors that corresponds to the number of members with their Survivor flag equal to TRUE.

We have set Crossover probability equal to 0.95 and Mutation probability 0.05.

Now in our modified GA we have introduced elitism by allowing the best 25 members, which were preserved in an extended array location, to take part in Crossover, but **not allowing** them to take part in Mutation so that their genetic content in the chromosome is not lost during the random Mutation process.

During the simulations we observed that improving SBL is the matter of major concern since for a very little improvement in SBL, SLL degrades heavily. So our stopping criterion for GA depends directly on the values of SLL, SBL and indirectly on the fitness of the best member. We may consider that our purpose of the project is served if SLL and SBL are very close to each other and both are very low. If we try to improve SBL beyond this point, SLL is bound to fall and vice versa. This restriction guided us to choose a point where we could stop. This fact can be clarified from the simulation results, given below.

## IV. SIMULATION RESULTS

The fitness function around which the whole GA method revolves is equation (5).

$$f^{(n)}(\chi) = w_1 . SLL^{(n)}(\chi)\big|_{f_0} + w_2 . SBL^{(n)}(\chi)\big|_{f_0 + m.prf}$$

Now, to have a particular value of the fitness function we have to choose proper weightings for SLL and SBL, i.e. we have to choose $w_1$ and $w_2$ in such a manner so that the resulting fitness function helps in suppressing both SLL and SBL through GA.

The different weight adjustments and the corresponding SLL and SBLs are shown in the table below:

TABLE 1: WEIGHTAGES AND CORRESPONDING SLL AND SBL

| Initial Population = 60, Generation = 2000 | | | |
|---|---|---|---|
| Probability of Crossover = 95% | | | |
| Probability of Mutation = 5% | | | |
| $w_1$ | $w_2$ | **SLL (dB)** | **SBL (dB)** |
| 1 | 1 | -42.3 | -21.4 |
| 1 | 3 | -34.8 | -24.2 |
| 1 | 6 | -29.3 | -28.6 |
| 1 | 10 | -24.7 | -31.0 |

It is evident from the above results that the weight ratio 1:6 is the optimal, for which the optimized sequence is shown in figure 1.

The optimal sequence gives SLL = -29.3 dB, SBL = -28.6 dB, which is a significant improvement over the previous SLL= -25.5 dB and SBL= -24.6 dB with binary time modulated linear array with only two logic states [3].

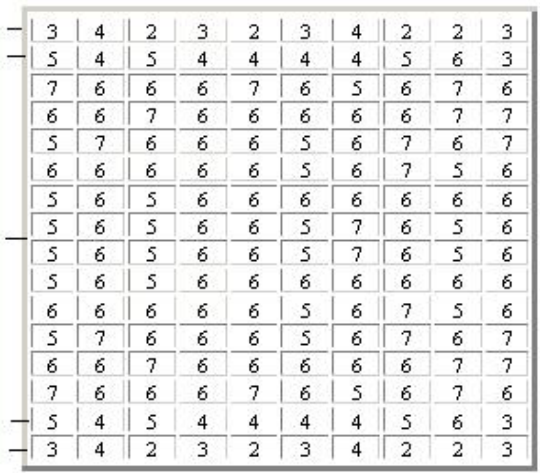

Figure 1: Optimized sequence of excitation, in y axis, the elements, in x axis, the time slots. Each number denotes the corresponding excitation level.

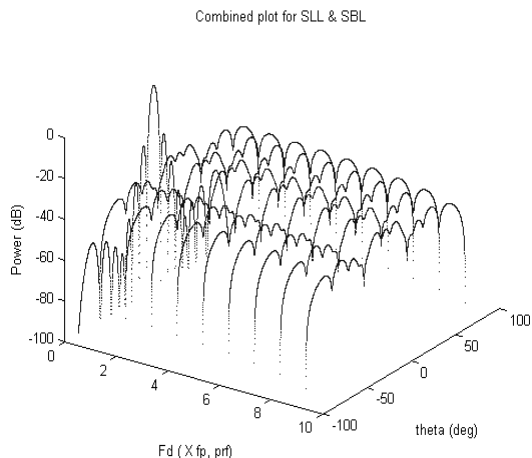

Figure 2: Power pattern for the optimal sequence

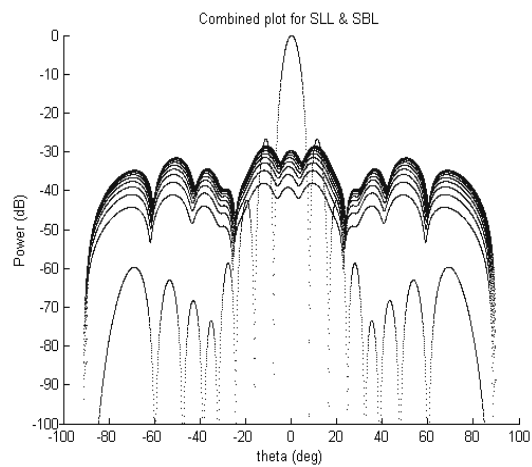

Figure 3: The optimized power pattern with respect to angle in different sidebands

But we want to mention that the results we have obtained may not be the best. We have used 2000 generations with population size of 60. If we could view the results from more number of generations then the optimized sequence might be better. But this kind of testing was beyond our scope as that would require a huge amount of time for simulation through personal computer with limited RAM capacity and CPU speed.

From our simulation in limited environment, further increase in population size does not improve the performance significantly, but increases the simulation time rapidly. During the simulations we observed that a very little improvement in SBL degrades SLL heavily. So if SBL requirements were not considered, SLL could have been improved significantly.

How the GA optimizes the SLL and SBL and maximizes fitness function are shown in Figure 5 and 6 respectively (all in absolute values).

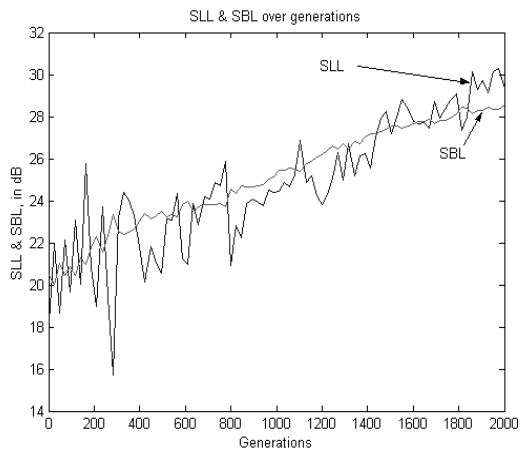

Figure 4: Plot of SLL and SBL over the 2000 generations.

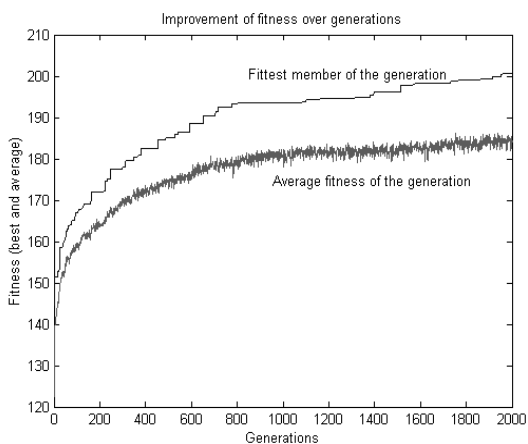

Figure 5: The upper plot shows the fitness of the best member and the lower plot shows the average fitness of the corresponding generation

The convergence rate of the GA can be obtained from the above plot (Fig 5), which depicts the variation of the average fitness value and the fitness value of the best member. From the nature of the curve, it can be seen that although the best fitness value is on the verge of saturation, the average value is still well below it, and it is obvious that many more generations would be needed for the average value to catch the value of the best member, i.e. the GA to saturate. But it would not be beneficial for us to wait till the saturation of GA because the value of the best member is on the verge of saturation and observing too many generations would not serve our purpose very well. So when SLL and SBL were very close to each other and both were very low, the GA was terminated to give the excitation pattern for the time and amplitude modulated linear array as was stated in the optimization procedure.

V.  ADVANTAGE OF TIME MODULATION

If we didn't use the time modulation here, to attain a Side lobe Level as low as −30dB, we had to excite the central element of the unmodulated array almost 32 times higher in magnitude than the other elements of the array. Thus it makes the amplitude taper of the array very high and those arrays are difficult to fabricate.

Whereas, here the maximum amplitude taper is 1:8 and by using the time modulation we get the −29dB low SLL. Though, the time modulated array has a sideband around the center frequency, which is absent in the unmodulated array, even if it is suppressed to −28dB and has very little effect on the operation of the array.

VI.  CONCLUSION

Here we have described a time modulated linear array with variable amplitude excitations to obtain low SLL and low SBL simultaneously. The time modulation period was divided into minimal time steps, and the variable excitation pattern or optimization sequence was obtained via a Genetic Algorithm on each array element.
Based on our simulation results SLL and SBL can be simultaneously suppressed below -30 dB and -29 dB respectively with this variable amplitude excitation. This is a considerable improvement over time modulated linear arrays with uniform amplitude excitations. Retaining 25 best members in each generation of the GA helps to retain all the best qualities in each generation even after mutation can change them. SLL and SBL can be made much better if the simulation is run for larger duration and with more number of generations.

We have also noticed that improving SBL is the most critical among the two and a little improvement in SBL heavily degrades the SLL of the array.

ACKNOWLEDGMENTS

We acknowledge the contribution of Dr. Bhaskar Gupta, Professor, Jadavpur University, Kolkata, who has given the idea of using time modulation in antenna arrays to make it more effective and state of the art.

REFERENCES

[1] Y. Rahamat Samii and E. Michielssen, Ed. *Electromagnetic Optimization by Genetic Algorithms*. John Wiley & Sons Inc, 1999, pp 1-27.
[2] David E. Goldberg, *Genetic Algorithm*, John Wiley & Sons, 2002, pp 1-22, 60-86.
[3] Shiwen Yang, Yeow Beng Gen, Anyong Qing, and Peng Khiang Tan, "Design of a Uniform Amplitude Time Modulated Linear Array With Optimized Time Sequences", *IEEE Transactions on Antenna and Propagation*, Vol. 53, No.7, July 2005, pp 2337-2339.
[4] Edward C. Jordan and Keith G. Balmain, *Electromagnetic Waves and Radiating Systems*. Prentice Hall of India Pvt. Ltd. 2003, pp 359- 374, 422-464.
**[5]** Balanis, *Antenna Theory: Analysis & Design*, 2nd Ed, John Wiley and Sons Inc, 1997, pp 249-328, 592-599.